\documentclass[conference]{IEEEtran}
\IEEEoverridecommandlockouts
\usepackage{cite}
\usepackage{amsmath,amssymb,amsfonts}
\usepackage{algorithmic}
\usepackage{graphicx}
\usepackage{textcomp}
\usepackage{xcolor}
\usepackage{todonotes}
\usepackage{booktabs}
\usepackage{algorithm}
\usepackage{multirow}
\usepackage{array}
\usepackage{dblfloatfix}
\usepackage{float}
\usepackage{cite}
\usepackage{subcaption}

\def\BibTeX{{\rm B\kern-.05em{\sc i\kern-.025em b}\kern-.08em
    T\kern-.1667em\lower.7ex\hbox{E}\kern-.125emX}}
\begin{document}

\title{Achieving Fairness in Dermatological Disease Diagnosis through Automatic Weight Adjusting Federated Learning and Personalization  \\
}

\author{\IEEEauthorblockN{Gelei Xu}
\IEEEauthorblockA{
\textit{Department of CSE} \\
\textit{University of Notre Dame}\\
IN, USA \\
gxu4@nd.edu}
\and
\IEEEauthorblockN{Yawen Wu}
\IEEEauthorblockA{\textit{Department of ECE} \\
\textit{University of Pittsburgh}\\
PA, USA \\
yawen.wu@pitt.edu}
\and
\IEEEauthorblockN{Jingtong Hu}
\IEEEauthorblockA{\textit{Department of ECE} \\
\textit{University of Pittsburgh}\\
PA, USA \\
jthu@pitt.edu}
\and
\IEEEauthorblockN{Yiyu Shi}
\IEEEauthorblockA{\textit{Department of CSE} \\
\textit{University of Notre Dame}\\
IN, USA \\
yshi4@nd.edu}

}

\maketitle

\begin{abstract}
Dermatological diseases pose a major threat to the global health, affecting almost one-third of the world’s population.  Various studies have demonstrated that early diagnosis and intervention are often critical to prognosis and outcome. To this end, the past decade has witnessed the rapid evolvement of deep learning based smartphone apps, which allow users to conveniently and timely identify issues that have emerged around their skins. In order to collect sufficient data needed by deep learning and at the same time protect patient privacy, federated learning is often used, where individual clients aggregate a global model while keeping datasets local. However, existing federated learning frameworks are mostly designed to optimize the overall performance, while common dermatological datasets are heavily imbalanced, e.g., the images of light skin types dominate those of dark skin types. This is due to the lack of medical professionals from marginalized communities, inadequate information about these communities, and socioeconomic barriers to participating in data collection and research. When applying federated learning to such datasets, significant disparities in diagnosis accuracy may occur. To address such a fairness issue, this paper proposes a fairness-aware federated learning framework for dermatological disease diagnosis. The framework is divided into two stages: In the first in-FL stage, clients with different skin types are trained in a federated learning process to construct a global model for all skin types. An automatic weight aggregator
is used in this process to assign higher weights to the client with higher loss, and the intensity of the aggregator is determined by the level of difference between losses. In the latter post-FL stage, each client fine-tune its personalized model based on the global model in the in-FL stage. To achieve better
fairness, models from different epochs are selected for each client to keep the accuracy difference of different skin types within 5\%. Experiments indicate that our proposed framework effectively improves both fairness and accuracy compared with the state-of-the-art.
\end{abstract}

\begin{IEEEkeywords}
Dermatological disease diagnosis, fairness, federated learning, fine-tuning
\end{IEEEkeywords}

\section{Introduction}
Dermatological disease is the fourth leading cause of non-fatal disease burden globally \cite{hollestein2014insight}. According to the estimation of National Institutes of Health (NIH), one out of five US citizens are under the risk of developing a debilitating dermatological problem in their lifetimes \cite{stern2010prevalence}. Dermatological patients can slow disease development with early intervention and dermatological surgery \cite{thomsen2020systematic}. In order to provide a timely and convenient diagnosis for dermatology patients, smartphone apps for dermatology diagnosing are widely put into use.
Deep learning models have shown great potential in automatic diagnosing within smartphone apps\cite{jones2018mobile,yousaf2020comprehensive,balapour2019mobile}. A large amount of data is necessary in the training of deep learning models to achieve high accuracy \cite{wu2020enabling}. However, a major challenge to enhance the performance of deep learning models for dermatology disease is the lack of large scale dataset. Data sharing constraint published by the Health Insurance Portability and Accountability Act (HIPAA)  make it impractical to centralize data from patients directly \cite{wu2021federated2}. For example, during the data collection of dermatological disease, patients take images through their device for a preliminary self-diagnosis and choose to upload those images at their own discretion \cite{velasco2019smartphone}. Normally, most of the patients are unwilling to share sensitive images with centralized server. Without an abundant dataset, it is infeasible to achieve satisfactory results through centralized training \cite{sun2016benchmark}.

Federated learning is a learning paradigm that enables collaboration between machine learning models while keeping data local for privacy \cite{wu2022distributed}. In a classical federated learning process, a federation of data owners (clients)  trains its model locally and uploads the training model to a centralized server. The server aggregates the local model and passes the global model back towards each client \cite{tan2022towards}. By leveraging federated learning, distributed dermatological images on the mobile device can be trained locally without uploading the training data directly to the server \cite{wu2021federated}.

Classical federated learning algorithm such as \textit{FedAvg} \cite{mcmahan2017communication} builds the centralized model with weight proportional to the data quantity of local clients. However, although federated learning may achieve high accuracy on average, the local accuracy of each client is not guaranteed. This fairness issue would be extremely prominent due to the skin-type bias in the existing dermatological dataset: For example, in one of the largest public dermatological dataset Fitzpatrick 17k \cite{groh2021evaluating}, most of the images are from lightest skin types and middle skin types, only 13.5\% images are from darkest skin types. The study \cite{kamulegeya2019using} pointed out significant racial disparity for Skin Image Search, an AI app that helps people identify skin conditions. It reports 70\% accuracy for the whole dataset, but only 17\% for dark skins. Compounding contributing factors for the heavy skin-type bias include a lack of medical professionals from marginalized communities, inadequate information about those communities, and socioeconomic barriers to participating in data collection and research. In the absence of a diverse population that reflects that of the U.S. population, potential safety or efficacy considerations could be missed. What is worse, with inadequate data, AI algorithms could misdiagnose people with brown and black skin, leading to increasing healthcare disparities  \cite{adamson2018machine}.

Several works have been focusing on improving fairness via federated learning. \cite{mohri2019agnostic} proposed \textit{AFL} framework to optimize the model for any target distribution formed by a mixture of client distributions. However, the model only maximizes the performance of the worst performing device and is only suitable when the number of clients is small.
\cite{li2019fair} designed \textit{q-FFL} to deal with the heterogeneous data distribution, which assigns higher weights to the model of poorly performed clients to reduce the accuracy variance of the simultaneously trained models. However, the fairness level is changing during training but the hyper-parameter $q$ is fixed, which leads to underfitting and overfitting problems. None of the existing works can successfully address the fairness issue in dermatological disease diagnosis. 

To address the skin-type bias issues of dermatological disease in federated learning systematically, we propose a framework with two stages: in-FL and post-FL. In the in-FL stage, a global model is trained by federated learning through clients of different skin types. To reduce the gap in the performance between each skin type as mentioned above, we present an algorithm which could adjust the weight of each client in model aggregation automatically by training loss. By adding weights to the client with larger training loss, the global model is always shifting its center towards the poorly behaved clients to decrease the performance gap, which improves the fairness among different skin types in each round. The post-FL stage allows each client to fine-tune its own personalized model based on the global model obtained in the in-FL stage. Conventionally, fine-tuning is used to adjust the model to reach higher accuracy. However, our result shows that fine-tuning can also be used as a powerful tool to achieve fairness. Fine-tuning the personalized model not only increases the accuracy of each skin type but also shortens the gap between easy-to-classify and hard-to-classify skin types.

In summary, the contributions of this paper are three-fold:
\begin{itemize}
\item We propose a framework with in-FL and post-FL to solve this fairness problem systematically. The in-FL stage uses weight adjusting federated learning to train a global fair model for clients of all skin types. The post-FL stage allows each client to train its personalized model to achieve better accuracy and fairness. This design not only takes full advantage of all the local data by aggregating a global model, but also considers local features of each skin type by fine-tuning a model for each client.
\item Our proposed framework achieves both good fairness and classification result. Experiment results on Fitzpatrick 17k demonstrate the effectiveness of our purposed method in comparison to the previous state-of-the-art methods.
\item Besides using the variation of accuracy directly as the metric of fairness, we also extend a binary multi-task fairness metric to multi-class. Compared with the variation metric, which can only be used to evaluate the fairness performance of the whole model, this metric also provides the fairness performance of each skin type.
\end{itemize}

\section{BACKGROUND AND RELATED WORK}
\subsection{Deep Learning in Dermatology}
Since most dermatological diseases are lesions with abnormal pattern and skin color, it is one of the most appropriate areas in medicine to use deep learning for automatic diagnosis \cite{goceri2021deep}. Recently, many deep learning frameworks are designed to be used in the diagnosis of dermatological diseases Melanoma. \cite{esteva2017dermatologist} proposed a dermatologist-level classification using deep neural networks in detecting Melanoma.  \cite{el2020deep} used a combination of Google-Net, ResNet101 and NasNet to differentiate Melanoma from benign nevus. In addition, researchers also use deep learning to achieve multi-disease classification. \cite{mishra2019interpreting} analyzed a convolution neural networks (CNN) based fine-grained classification process for ten common East Asian dermatological conditions and reached an accuracy of more than 85\%.  \cite{goceri2021deep} used a pretrained DenseNet201 with a specified loss function to classify five common facial dermatological diseases.    \cite{srinivasu2021classification} developed a mobile app based on MobileNetV2 and Long Short Term Memory (LSTM), achieving an accuracy of 85.34\% on HAM10000 dataset.  

Most of the existing dermatological diagnosis frameworks are designed for centralized learning. However, in clinical settings, dermatological images are mostly scattered on patients' mobile devices and it is difficult and costly to send data directly to the central server due to privacy issues. Therefore, an approach learning a shared model from decentralized data is needed.

\subsection{Federated Learning in Dermatological Disease}
Federated learning allows data sharing among multiple decentralized edge devices while keeping raw data locally on devices, which is in contrast to the traditional machine learning techniques where datasets are uploaded directly to the server \cite{konevcny2015federated}. Federated learning provides access to heterogeneous data and is preferred in use cases where data security and privacy are key concerns \cite{mothukuri2021survey}.
In a typical federated learning algorithm \textit{FedAvg} \cite{mcmahan2017communication}, the asynchronous update is performed in the communication round by round. In each round $t$, the central server activates a fraction of clients $C^t$ and sends them the latest global model. Each client $c \in C^t$ trains for several epochs locally on local dataset $D_c$ to update their parameter $\theta^{t+1}_c$ by minimizing the loss $\mathcal{L}$. Then the server aggregates the local parameters to form a global model through weight averaging $\theta^{t+1} \leftarrow \sum_{c \in C^t}\frac{|D_c|}{\sum_{i \in C^t}|D_i|}\theta^{t+1}_c$. This learning process repeats until the model converges. 

Federated learning is used for privacy protection in dermatological disease diagnosis. \cite{agbley2021multimodal} presented a federated learning model which fuses two modalities in diagnosing Melanoma disease and obtained a similar F1 and accuracy score with the centralized model.  \cite{wu2021federated} proposed a federated contrastive learning framework for dermatological diagnosis and achieved a higher recall and precision than classical federated learning methods.  \cite{hossen2022federated} suggested a CNN model for four types of human skin disease classification which showed a higher recall and precision than several benchmark CNN algorithms. 

Existing works have only focused on the classification accuracy in dermatological disease diagnosis. However, the fairness issue has been ignored between different skin types. If only overall accuracy is considered in dermatological disease diagnosis, since most of the data samples are light skin, the accuracy of light skin will be significantly larger than dark skin type, which leads to increasing healthcare disparities. Therefore, an approach based on federated learning to address the fairness problem caused by skin-type bias is needed. 
\subsection{Fairness in Federated Learning}
Ensuring fairness among users has been an important topic since it determines the willingness of the participation of clients and influences the effect of the model. According to \cite{shi2021survey}, existing studies of fairness in federated learning can be classified into four aspects: client selection, model optimization, contribution evaluation and incentive mechanism. As this work focuses on model optimization, only works in this area are reviewed here. 
 \cite{mohri2019agnostic} put forward a framework \textit{AFL} where the centralized model is optimized for any unknown testing data distribution. They stimulate the testing distribution as the mixture of the data distribution of clients, and the weight of each client is determined by the proportion of its data quantity in the entire training data. This does not affect the model performance of other clients as long as they do not increase the loss of the client with poor performance. However, in their formulation, only the performance of the worst client is optimized so this algorithm only performs well when the number of clients is small. To make up for the scalability of \textit{AFL},  \cite{li2019fair} proposed \textit{q-FFL} which reaches a more uniform accuracy distribution. The fairness degree can be adjusted by choosing different hyper-parameter $q$ to assign higher weight to the client with poor performance. However, \textit{q-FFL} is unable to adjust the weight flexibly during the middle of training, which leads to underfitting and overfitting problems when the degree of fairness is changing in the training process. Therefore, a weight adjusting framework to assign weights based on the performance of clients automatically is needed.

\begin{figure*}[ht]
  
  \centering
  \includegraphics[width= \linewidth]{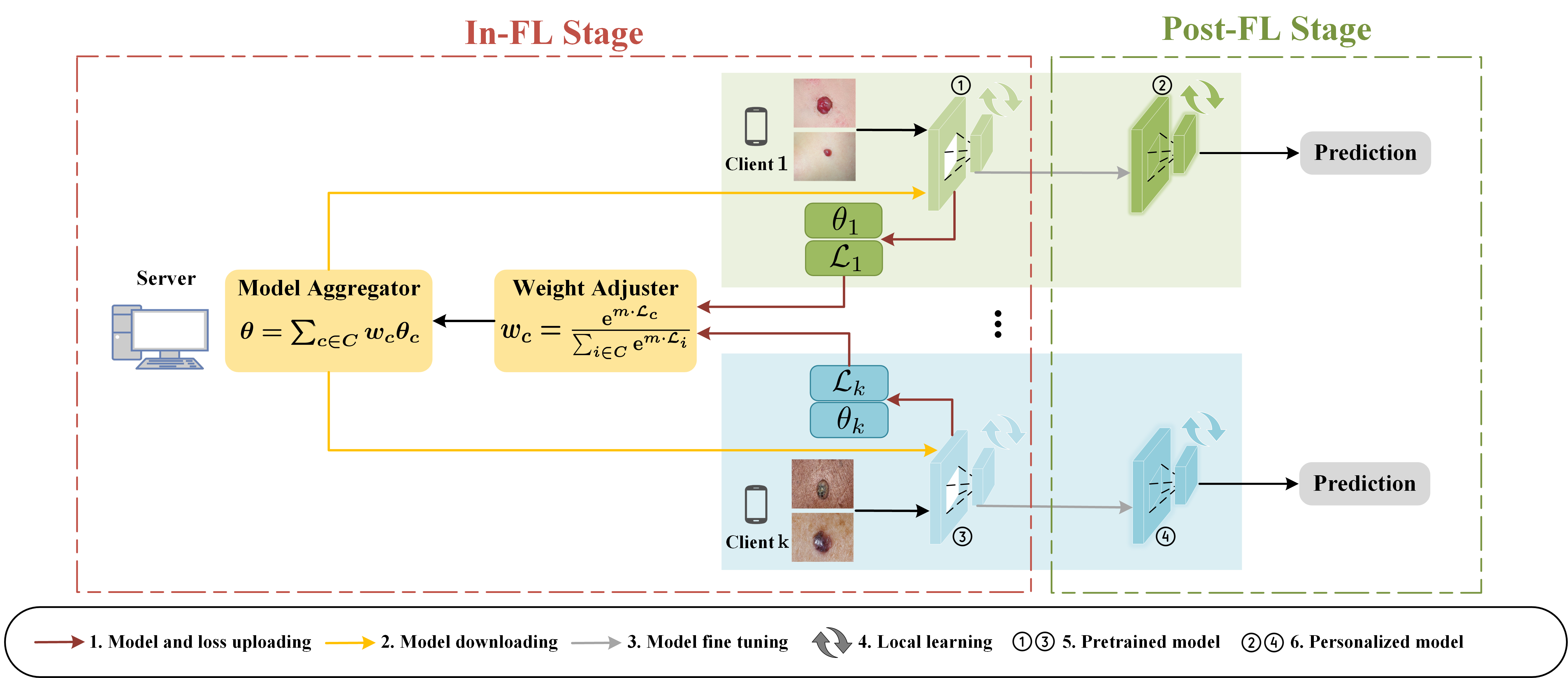}
  \caption{The framework for dermatological diagnosis based on federated learning}
  \label{framework}
\end{figure*}

\section{OVERVIEW OF PROPOSED FRAMEWORK}
The overview of our dermatological disease classification framework is shown in Fig.\ref{framework}. The framework is divided into two stages. In the first in-FL stage, federated learning is performed to train a globalized model, in which six clients are used in total and each client represents one skin type. The main difference between our method and typical federated learning is the weight adjuster in the server we used in Fig.\ref{framework}. For every global epoch, each client trains its model for several local epochs and then uploads its model parameter and training loss towards the weight adjuster. Weight adjuster signs weights to each client. Client with higher loss will be assigned with higher weight through the exponential function. Then parameters and weights are transferred to model aggregator, in which the updated model parameter is computed by the weighted sum of the model parameter of each client. After model aggregation, each client downloads the updated model and starts a new epoch. In the post-FL stage, clients will be trained locally for several epochs using the global model obtained in the in-FL stage to fine-tune its personalized model. In the end, each skin type has its own personalized model to make predictions.

To enhance the fairness of the framework, two specific techniques are used. First is the automatic selection of the scaling factor $m$ in the in-FL stage. In weight adjuster, in order to adjust the weights between models in time to achieve maximum fairness, $m$ is automatically assigned by the range of loss of each client. When the range is larger, $m$ will increase to add more weights to the poorly performed model, and vice versa. This design effectively avoids the underfitting and overfitting of fairness caused by improper weight assigning. Second is the model selection of different epochs in the post-FL stage. In personalized training, the accuracy of each client will vary greatly in a same epoch due to the different training speeds of each skin type. At the same time, it may also appear that one model converges with much higher accuracy than the other. To limit the accuracy of different clients in a fixed range for more fairness, we adjust the epochs from which the models are selected for different clients.

\section{Methodology}
\subsection{Weight Adjusting in In-FL}\label{sect:fl_weighting}
Recently, most federated learning paradigms use \textit{FedAvg} as the backbone, which weights models according to the data quantity of each client. This algorithm makes sense in some cases since a client can be as small as an individual mobile device to as large as an entire organization, and the client with a larger amount of data deserves more weight. But obviously, this weighting method does not apply to our dermatological fairness problem since all skin types are equivalent clinically and both accuracy and fairness should be considered. Therefore, a fairness-aware federated learning framework is needed. However, existing federated learning frameworks are not strong enough to satisfy the extremely imbalanced dataset in dermatological disease even if weight adjusting is implemented in their training process. Therefore, it is necessary to propose an automatic weight assigning method to achieve fairness in the dermatological dataset.

A natural idea of achieving fairness is to assign more weight to the client with poor performance. Intuitively, the performance of each client can be measured by the loss value of the model. The average loss $\mathcal{L}_c$ for client $c$ is calculated as follows:
$$
\mathcal{L}_c = \frac{1}{N_c} \sum_{n=1}^{N_c} l_{ce}(p_n, t_n)
$$
where $N_c$ is the amount of data sample in client $c$, $p_n$ is the vector of prediction after $Softmax$ activation, $t_n$ is the one-hot target vector of $n$-th data. Here we take cross entropy loss $l_{ce}$ as our loss function:
$$
l_{ce}(p_n, t_n)=-\sum_{i=1}^{M}t_n^{(i)}\log(p_n^{(i)})
$$
where $M$ is the number of classes.

Based on the loss function of each client $\mathcal{L}_c$, we construct our weight adjusting function as shown below:
$$
{w_c} = \frac{{\rm e}^{m\cdot\mathcal{L}_c}}{\sum_{i\in C}{\rm e}^{m\cdot\mathcal{L}_i}}
$$
The design of the function is based on the following three ideas:

\begin{itemize}

\item In practice, the loss value between models are close to each other, simply linearly assigning the loss as the weight of each client is not enough to achieve fairness. Therefore, we wrap the loss value with an exponential function to map loss to the weight of the model and amplify the difference between loss values.

\item In order to make our algorithm more flexible and efficient, we aim to increase the power of weight adjustment when the difference between the loss values is large. Therefore, we add a scaling factor $m$ toward the loss value. $m$ is an integer with an initial value of 1 and range$(1, \mathcal{M})$, where $\mathcal{M}$ is the upper bound of $m$. Experiments of different $\mathcal{M}$ value is shown in Table \ref{parameter} in section \ref{experiment}. $m$ gets larger when $L_{max}$ is larger than $\mathcal{Q}\times L_{min}$, where $L_{max}$ and $L_{min}$ are the maximum and minimum loss value of all clients, $\mathcal{Q}$ is a hyper-parameter. 

\item To ensure the consistency of weights during each round of training, the weight adjusting function is normalized.

\end{itemize}

By aggregating the model with the weights above, we can calculate the updated model by:

\begin{equation}
    \theta = \sum_{c \in C}w_c\theta_c
\end{equation}

After model aggregation, clients download the new global model and start a new round. The detailed procedure of in-FL is shown in Algorithm \ref{algorithm}.

\begin{algorithm}
\caption{Fair Federated Learning with Automatic Weight Adjusting (\emph{FedAuto})}
\label{algorithm}
\begin{algorithmic} 
\STATE \hskip\dimexpr-\algorithmicindent \textbf{Input:} $C, T, E, B, \theta, \mathcal{Q}, \mathcal{M}$ 
\STATE $m \leftarrow 1$
\FOR{each round $t = 1, 2, 3...T$}
\STATE $L$ is an empty set
  \FOR{each client $c \in C$ in parallel}
      \STATE $\theta_c \leftarrow \theta$\hfill $\triangleright$ \textit{download global model}
      \STATE $\theta_{c}, \mathcal{L}_c \leftarrow $LocalUpdate$(c, \theta_c)$
      \STATE $L$ append $\mathcal{L}_{c}$
  \ENDFOR
  \IF{$L_{max} > \mathcal{Q} \times L_{min}$ and $m \leq \mathcal{M}$} 
  \STATE $m \leftarrow m+1$
  \ENDIF
  \FOR{each client $c \in C$}
  \STATE ${w_{c}} \leftarrow \frac{{\rm e}^{m\cdot\mathcal{L}_{c}}}{\sum_{i\in C}{\rm e}^{m\cdot\mathcal{L}_{i}}}$\hfill $\triangleright$ \textit{adjust weight}
  \ENDFOR
  \STATE $\theta \leftarrow \sum_{c \in C}w_c\theta_c $\hfill $\triangleright$ \textit{aggregate local models}
\ENDFOR

\STATE \hskip\dimexpr-\algorithmicindent\textbf{LocalUpdate($k, \theta_k$):}\hfill $\triangleright$ \textit{run on client $k$}

\STATE $\mathcal{B}\leftarrow$ (split dataset of $k$ into batches of size $B$)
\FOR{each local epoch $i$ from $1$ to $E$}
  \FOR{batch $b\in \mathcal{B}$}
      \STATE {$\theta_k \leftarrow \theta_k - \eta\nabla\mathcal{L}(\theta_k;b)$}
  \ENDFOR
\ENDFOR
\STATE return $\theta_k, \mathcal{L}_k$ to server
\end{algorithmic}
\end{algorithm}

\subsection{Fine-Tuning in Post-FL}
In our framework, we use fine-tuning to train the global model on each client separately to obtain the personalized model of each skin type. Conventionally, fine-tuning is believed to be more efficient than training from scratch. In our framework, however, in addition to efficiency, it can also provide the following benefits:

Firstly, fine-tuning can enhance local features of each skin type, which improves the personalization performance. In-FL training in Section \ref{sect:fl_weighting} generates a global and generalized model for all six skin types, which is based on the global pathological features while avoiding local features of each skin type. However, since the distribution of dermatological diseases differs among different skin types \cite{adelekun2021skin}, the global model may not have a good accuracy on all clients.
Local features is powerful in enhancing the global model
for better classification results for each client. By combining the newly-learned local skin type features with the global and generalized dermatological features, each client can effectively improve its own personalized model, which also improves the overall diagnosis performance of all clients.

Secondly, fine-tuning can shorten the performance gap between each skin type and improves the fairness across different skin types. As mentioned in \cite{toneva2018empirical}, 
different samples have intrinsically different levels of difficulties for learning.
Some samples are hard to classify and hard to learn by the model, which means that these samples need more train iterations. Usually, unforgettable samples are easily recognized and contain more obvious class attributes. Since the skin color of the pathological region is an important reference for clinical diagnosis of dermatological disease \cite{habif2017skin}, the difficulties of classification may vary due to the properties of different skin types. 
Therefore, it is necessary to shorten the gap between easy-to-classify skin types and hard-to-classify skin types. 
This goal can be achieved by performing the personalized model selection independently for different clients.
In the personalized model selection, different clients can obtain the best model in different epochs during fine-tuning due to the reasons below: 
First of all, it alleviates the fairness problem caused by different convergence speeds by training hard-classified skin types with more epochs to obtain similar results. Besides, it can also mitigate the fairness problem caused by the large accuracy gap after convergence. For skin types that converge with higher accuracy, we select models in the middle of their training process where accuracy is close to hard-classified skin types to achieve the trade-off between accuracy and fairness.

\section{Experiments}\label{experiment}

\noindent\textbf{Dataset.} The proposed method is evaluated on Fitzpatrick 17k dataset, which contains 16577 images in total with nine different skin conditions \cite{groh2021evaluating}. The skin type label is based on Fitzpatrick scoring system. Since our experiments are conducted mainly on skin types, we remove the images with Fitzpatrick label -1, which means the skin type is unknown. The rest of the images are labeled with six Fitzpatrick skin types from 1 to 6, where a smaller label represents lighter skin and a larger label represents darker skin. Detailed information of the skin type distribution of Fitzpatrick 17k is as shown in Table \ref{dataset}. There are significantly more images of light skin than dark skin, skin type 6 even only accounts for 4\% of the total six skin types. 
Such an imbalanced skin type distribution is a challenging and effective setting to evaluate our fairness federated learning methods.

\begin{table}[ht]
  \caption{Skin Type Distribution of Fitzpatrick 17k}
  \label{dataset}
  \begin{tabular}{cccccccc}
    \toprule
    Skin Type & 1&2&3& 4&5&6&Total\\
    \midrule
    Number&2944 & 4807 & 3306 &2781 &1531 &634&16003\\

  \bottomrule
\end{tabular}
\end{table}

\noindent\textbf{Preprocessing details.} The dataset is randomly separated into a training set of 60\% data, a validation set of 20\% data, and a test set of 20\% data, respectively. 
The test set is not used in any stage of in-FL or post-FL. All input images are resized to 128$\times$128. Horizontal flipping, vertical flipping, rotation, scaling, and autoaugment are used to augment the data \cite{cubuk2018autoaugment}. Resampling is used to maintain the balance between classes.

\noindent\textbf{Federated setting and training details.} Six clients are used in this federated framework and each contains images of one skin type. All clients are selected and updated in each round. The in-FL stage is performed for 100 communication rounds. In each round, each client is activated and trained for 5 local epochs. In the weight adjusting method, the hyper-parameter $\mathcal{M}$  is set to 3 and $\mathcal{Q}$ is set to $\frac{3}{2}$. VGG-11 is used as the backbone. Adam optimizer is used with the batch size of 128 and the initial learning rate is 1e-4 with a cosine decay. The post-FL stage has the same parameters as the in-FL stage. In post-FL, each client is trained on its own dataset for 100 epochs to get the personalized model. Both in-FL and post-FL stages are performed on one Nvidia V100 GPU.

\begin{table*}[ht]
  \centering
  \caption{Result of Weight Adjusting of in-FL and Baselines }
  \label{in-FL}
  \begin{tabular}{ccccccccccccc}
    \toprule
    Methods & Accuracy & Precision & Recall & F1  & Variation & Type 1 &Type 2&Type 3&Type 4&Type 5&Type 6\\
    \midrule
    \textit{FedAvg } \cite{mcmahan2017communication}& 0.615& 0.547 & 0.674 & 0.582&0.00461 & 0.611 & 0.612 & \textbf{0.658} & 0.653 & 0.535 & 0.467 \\
   \textit{FedEqual} & 0.617 &0.556&0.681&0.589& 0.00395 & 0.628 & 0.615 & 0.649 & 0.618 & 0.600 & 0.458\\
    \textit{q-FFL } \cite{li2019fair}& 0.625& 0.622 & 0.684 & 0.642 & 0.00336 & 0.623 & 0.635 & 0.646 & 0.645 & 0.576 & 0.483\\
    \textit{FedLoss} & 0.631& 0.627 & 0.682& 0.644 & 0.00153 & 0.617 & 0.632 & 0.649 & 0.647 & 0.614 & 0.533\\
    \textit{FedExp} & 0.632& 0.631& \textbf{0.696} & 0.651&  0.00122 & 0.622 & 0.633 & 0.629 & \textbf{0.663} & \textbf{0.640} & 0.550\\
    \textit{FedAuto }(ours) & \textbf{0.643}& \textbf{0.643} & 0.694 & \textbf{0.660} &  \textbf{0.00065} & \textbf{0.634} & \textbf{0.648} & 0.657 & 0.649 & 0.631 &\textbf{0.575} \\
    \bottomrule
  \end{tabular}
\end{table*}

\noindent\textbf{Baselines.} We compared our proposed baseline with five baselines. \textit{FedAvg} is a standard FL aggregation method and it uses the number of samples of each client as the weight for model aggregation. \textit{FedEqual} sets all weights of clients equally. \textit{FedLoss} uses loss and \textit{FedExp} uses exponential loss as weight. 
We also compare our method with \textit{q-FFL}, a state-of-the-art method in federated learning for achieving fairness, we explored its performance with different values of hyper-parameter $q$ = 0, 1, 5 and report the average performance.

\noindent\textbf{Fairness metrics.} Two metrics are used in this paper to evaluate fairness:
\begin{itemize}
    \item Since the goal of fairness is to achieve accuracy parity by measuring the uniformity across federated learning devices \cite{li2019fair}, the variance of accuracy can be used to measure the degree of uniformity between each client \cite{shi2021survey}, as shown below:

\begin{equation}
    Var = \frac{\sum_{c\in C}( Acc_c - \overline{Acc})^2}{|C|}
\end{equation}

where $Acc_c$ is the accuracy of client $c$, $\overline{Acc}$ is the average accuracy of all clients.

    \item We also quantify fairness by extending the gap and worst metric in \cite{ramapuram2021evaluating} to multi-class classification. For each skin type $s$, $Acc(s)$  denotes the accuracy of $s$, $ Acc(\neg s)$ denotes the accuracy belongs to other skin types except $s$. Define:

\begin{equation}
    Acc^{gap}(s) = |Acc(s) - Acc( \neg s)|
\end{equation}

\begin{equation}
    Acc^{worst}( s) = \min(Acc(s) , Acc(\neg s))
\end{equation}

 Obviously, a fair model minimizes the accuracy difference and maximizes the worst accuracy between each skin type to achieve fairness as well as maintain accuracy. Therefore, skin type $s$ with lower $ Acc^{gap}(s)$ and higher $Acc^{worst}(s)$ will contributes more towards fairness. To evaluate the fairness of a whole model, we can simply take $\frac{\sum_{s\in S}Acc^{(gap|worst)}(s)}{|S|}$, where $S$ is the set of all skin types.
\end{itemize}

\noindent\textbf{Classification performance metrics.} Following the previous dermatological classification works \cite{back2021robust,wu2022fairprune,morgado2021incremental}, this paper uses four evaluation metrics to assess the classification performance of our model: Accuracy, Precision, Recall and F1 score. Accuracy indicates the number of dermatological images which is correctly classified. Recall is an important evaluation metric in medical diagnosis domain, which denotes the number of images for each class that is correctly classified. Precision represents the ability of the classifier not to classify a negative sample as positive. Usually, it is difficult to achieve both high precision and recall, therefore F1-score is needed to evaluate the trade-off between these two metrics.
The formulas of these four metrics are shown below:
\begin{equation}
    Accuracy = \frac{TP+TN}{TP+TN+FP+FN}
\end{equation}

\begin{equation}
    Precision = \frac{TP}{TP+FP}
\end{equation}

\begin{equation}
    Recall = \frac{TP}{TP+FN}
\end{equation}

\begin{equation}
    F1 = \frac{2\times Precision \times Recall}{Precision + Recall}
\end{equation}
where $TP$, $TN$, $FP$ and $FN$ indicate true positive, true negative, false positive and false negative, respectively. Metrics of the whole dataset are the average value weighted by the number of true instances for each class.

\subsection{Result on In-FL}

The result of the in-FL stage is shown in Table \ref{in-FL}, where accuracy, precision, recall and F1 are reported as classification metrics and the variance of accuracy is reported as fairness metric. 
Our automatic weight adjusting method achieves the smallest variance, which is more than 7 times smaller than \textit{FedAvg} method. At the same time, our method also reaches the highest accuracy, precision and F1. It outperforms \textit{FedAvg}, \textit{FedEqual}, \textit{q-EF}L, \textit{FedLoss}, \textit{FedExp} 2.8\%, 2.6\%, 1.8\%, 1.2\%, 1.1\% for accuracy, and 7.8\%, 7.1\%, 1.8\%, 1.6\%, 0.9\% for F1. In general, fairness is often seen as a trade-off with classification performance, but results show that our proposed method not only achieves the fairness, but also increases the overall classification performance.

\textit{FedAvg} and \textit{q-FFL} perform the model aggregation by using the number of samples as the weight. However, the results show that weighting by the number of samples is not a good practice for problems such as skin disease diagnosis due to the extreme imbalance skin types of the dataset, which results in poor fairness. What's more, comparing the accuracy of six skin types, we find that the accuracy of skin type 1-4 is generally maintained at a uniform benchmark, while the accuracy of skin type 5 can be increased significantly by loss adjusting methods \textit{FedLoss}, \textit{FedExp} and \textit{FedAuto}. However, a huge accuracy gap still remains between skin type 6 and the other five skin types even in \textit{FedAuto} method. Therefore, it is necessary to use post-FL to achieve higher accuracy and further fairness.

\begin{table}[ht]
\centering
\caption{ Result of different fixed scaling factor $m$ and upper bound $\mathcal{M}$ }
\label{parameter}
\begin{tabular}{cccc}
\toprule
Methods         & Accuracy  & F1    &Variation\\ \midrule
Fixed $m=2$     & 0.635       & 0.645 &0.00123\\
Fixed $m=3$     & 0.640       & 0.653 &0.00096\\
Fixed $m=4$     & 0.622       & 0.639 &0.00182\\ \midrule
$Auto$ with $\mathcal{M}=2$ & 0.638      & 0.657 &0.00078\\
$Auto$ with $\mathcal{M}=3$ & \textbf{0.643}       & \textbf{0.660} & \textbf{0.00065}\\
$Auto$ with $\mathcal{M}=4$ & 0.630      & 0.658  & 0.00180 \\ \bottomrule
\end{tabular}
\end{table}

\begin{table*}[ht]
\centering
\caption{ Result of fine-tuning of post-FL and Baselines}
\label{post-FL}
\begin{tabular}{cccccccccc}
\toprule
&   & Methods    & Type 1 & Type 2 & Type 3 & Type 4 & Type 5 & Type 6 & All    \\ \midrule              
\multicolumn{2}{l}{\multirow{4}{*}{Gap$\downarrow$}}   & \textit{FedAvg} &0.0395 &  \textbf{0.0082}  & 0.0589 &0.0281 & 0.0406   &0.1293  &0.0508        \\ 
\multicolumn{2}{l}{} & \textit{q-FFL}   & 0.0509  &0.0521 & \textbf{0.0093} & 0.0309 & 0.0381 & 0.0974   & 0.0465       \\

\multicolumn{2}{l}{} & \textit{FedLoss}  & 0.0445  &0.0294& 0.0232  &\textbf{0.0225}  & 0.0345 &0.0514        &  0.0344      \\
\multicolumn{2}{l}{} & \textit{FedAuto} & \textbf{0.0308} &0.0101  & 0.0127 & 0.0447 & \textbf{0.0330} & \textbf{0.0115} & \textbf{0.0238} \\ \midrule
\multicolumn{2}{l}{\multirow{4}{*}{Worst$\uparrow$}} & \textit{FedAvg} & 0.6620&0.6918  & 0.6821    & 0.6710       &0.6904  & 0.5700  &  0.6612  \\
\multicolumn{2}{l}{}  & \textit{q-FFL}  & 0.6620 & 0.6879  &\textbf{0.7016} &0.6780  & 0.6999       & 0.6100       & 0.6732       \\

\multicolumn{2}{l}{} & \textit{FedLoss}  & 0.6730 &0.7005 & 0.6910 & \textbf{0.7055} & 0.7060 & 0.6600       & 0.6893       \\
\multicolumn{2}{l}{} & \textit{FedAuto}  &\textbf{0.6860}        & \textbf{0.7040}       & 0.7010       & 0.7033       & \textbf{0.7079}       & \textbf{0.7000}       & \textbf{0.7003 }      \\ 
\midrule
\end{tabular}
\end{table*}

Next, we evaluate the effectiveness of our dynamic weighting methods to improve the FL fairness.
Vanilla fixed weighting method achieves a lower fairness than our methods even at the cost of extensive hyper-parameter tuning.
Table \ref{parameter} shows the result of different fixed scaling factor $m$ and upper bound $\mathcal{M}$. When $m$ is fixed, $m$ value in the weight adjusting function will not be changed during the whole training process. Accuracy, F1 score and variation are reported in the table. The result shows that $\mathcal{M}=3$ reaches the highest accuracy, F1 score and the lowest variation, which means that $\mathcal{M}=3$ is the best upper bound setting for the weight adjusting function. The result of $\mathcal{M}=2$ is close to $\mathcal{M}=3$, but the result of $\mathcal{M}=4$ drops suddenly, this is because the difference in weights mapped by the exponential function widens rapidly as the value become larger. We can also see from Table \ref{parameter} that adjusting the intensity of weight automatically performs better than fixed $m$, which validates the automatic weight adjusting methods.

\subsection{Result on Post-FL}

Next, we evaluate the effectiveness of our post-FL for achieving fairness.
We compare the fairness performance of \textit{FedAvg}, \textit{q-FFL}, \textit{FedLoss}, \textit{FedAuto} using gap and worst metrics after fine-tuning in post-FL stage, the result is shown in Table \ref{post-FL}. Firstly, comparing Table \ref{post-FL}  with Table \ref{in-FL}, we can see that for every method, fine-tuning greatly improves the accuracy of all six skin types since the worst value outperforms its corresponding accuracy in Table \ref{in-FL}. Secondly, for skin type 6, the accuracy improvement of \textit{FedLoss} and \textit{FedAuto} is significantly larger than \textit{FedAvg} and \textit{q-FFL},  which demonstrates the effectiveness of weight adjustment by loss. Thirdly, we can derive from the result of \textit{FedAuto} and \textit{FedLoss} that the advantages of weight adjusting have not been eliminated, and are more prominent in skin types 5-6 with fewer data. All in all, our proposed \textit{FedAuto} method boosts the accuracy of skin type 6 to the same level with the other skin types without reducing the performance of the other skin types, which greatly improves fairness.

\begin{figure}[ht]
  
  \centering
  \includegraphics[width=\linewidth]{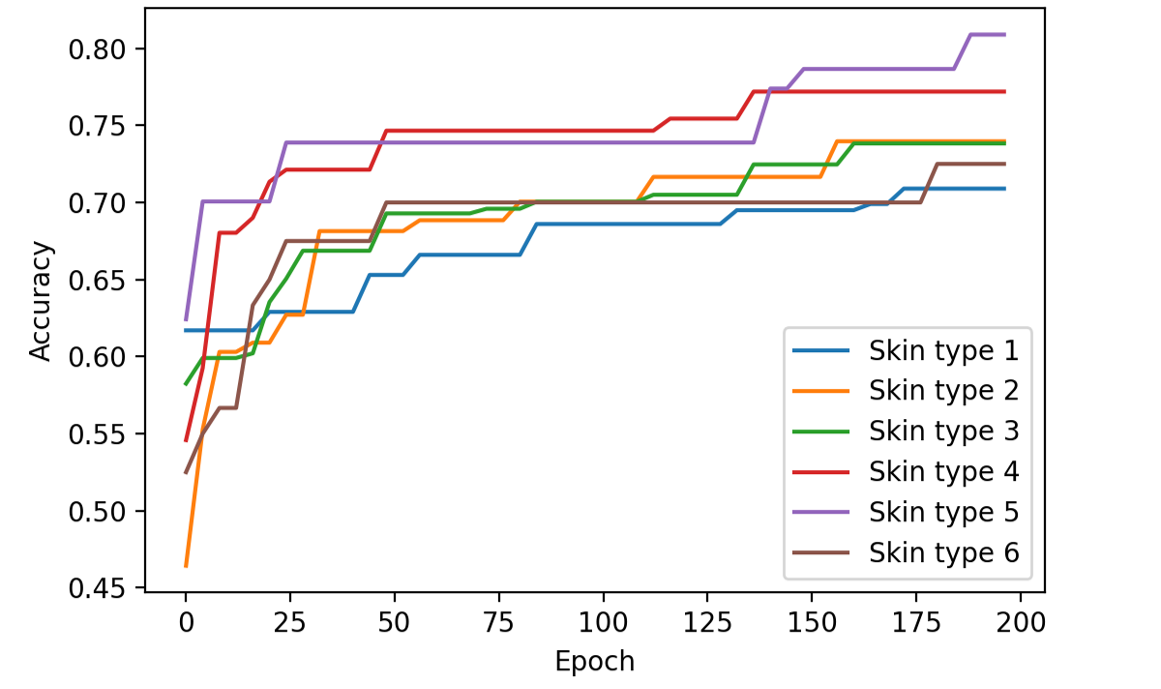}
  \caption{The training process of each skin type}
  \label{epoch}
\end{figure}

After confirming the effectiveness of \textit{FedAuto}, we run extra 100 epochs to select a personalized model for each skin type to reach the best trade-off between fairness and accuracy. The result of accuracy versus epoch of each skin type is shown in Fig.~\ref{epoch}. It can be seen that the training speed and accuracy of each skin type vary greatly. For example, the accuracy of skin type 5 exceeds 0.7 in the first 25 epochs and reaches a high accuracy towards 0.8 at the end of the training process. However, skin type 1 has a smooth training curve and finally arrives at 0.7 at the end of training. Therefore, to achieve both fairness and high accuracy, a specified model selection for each skin type is needed. After analyzing Fig.~\ref{epoch}, we select the model for each skin type with an accuracy between 0.7 and 0.75 to keep the accuracy gap within 0.05 to maintain fairness. The detailed information of the training and selected model is shown in Table \ref{training}. We also compute the overall accuracy and the variation of accuracy using our fine-tuned model. The result shows that the model after the post-FL stage not only outperforms the accuracy of the original model by 9.6\%, but also compresses the variance of accuracy to only 32\% of the original model.

\begin{table}[ht]
\caption{ Result of the training process of each skin type}
\label{training}
\centering
\begin{tabular}{cccc}
\hline
\multicolumn{1}{c}{Skin Type} &
  \begin{tabular}[c]{@{}c@{}}Best Model\\  Accuracy\end{tabular} &
  \begin{tabular}[c]{@{}c@{}}Selected Model\\  Accuracy\end{tabular} &
  \begin{tabular}[c]{@{}c@{}}Select Model \\ Epoch\end{tabular} \\ \hline
1                                     & 0.709 & 0.709    & 175                  \\
2                                     & 0.740 & 0.740    & 160                  \\
3                                     & 0.738 & 0.738    & 166                  \\
4                                     & 0.772 & 0.746    & 52                   \\
5                                     & 0.809 & 0.738    & 27                   \\
6                                     & 0.725 & 0.725    & 184                  \\ \hline
\multicolumn{1}{c}{Total Accuracy}    &       & 0.734    & \multicolumn{1}{c}{} \\
\multicolumn{1}{c}{Accuracy Varation} &       & 0.00015& \multicolumn{1}{c}{} \\ \hline
\end{tabular}
\end{table}

\section{CONCLUSION}
This paper proposes a framework based on federated learning to deal with skin-type bias problems in dermatological disease diagnosis. The framework is constructed in two stages. In the first in-FL stage, the model achieves fairness by putting more weight on clients with higher losses. In the second post-FL stage, each skin type fine-tunes its own personalized model based on the global model obtained in the first stage, which results in higher accuracy and fairness. The experimental results on Fitzpatrick 17k dataset of different skin types show the effectiveness of our proposed approaches for reducing skin-type bias in dermatological disease diagnosis.

\bibliographystyle{IEEEtran}
\bibliography{main}

\end{document}